\documentclass[11pt,a4paper]{article}
\usepackage[hyperref]{emnlp-ijcnlp-2019}
\usepackage{times}
\usepackage{latexsym}

\usepackage{url}

\aclfinalcopy % Uncomment this line for the final submission

\usepackage{color,soul}
\usepackage{amsmath}
\usepackage{amsfonts}
\usepackage[capitalize]{cleveref}
\usepackage{MnSymbol}
\usepackage{graphicx}   
\usepackage{subcaption}

\DeclareMathOperator*{\ReLU}{\text{ReLU}}
\DeclareMathOperator*{\softmax}{\text{softmax}}
\DeclareMathOperator*{\argmax}{\text{argmax}}
\DeclareMathOperator*{\softplus}{\text{softplus}}
\DeclareMathOperator*{\KL}{\text{KL}}

\title{Variational Fusion for Multimodal Sentiment Analysis}

\author{Navonil Majumder$^{\dagger}$, Soujanya Poria$^{\ddagger}$, Gangeshwar Krishnamurthy$^{\Phi}$,\\\textbf{Niyati Chhaya$^{\nabla}$, Rada Mihalcea$^{\lVert}$, Alexander Gelbukh$^{\dagger}$}\\
$^{\dagger}$Centro de Investigaci\'on en Computaci\'on, Instituto Polit\'ecnico Nacional, Mexico\\
$^{\ddagger}$Information Systems Technology and Design, SUTD, Singapore \\ 
$^{\Phi}$A*STAR AI Initiative, Institute of High Performance Computing, Singapore \\
$^{\nabla}$Adobe Research, India\\
$^{\lVert}$Computer Science \& Engineering, University of Michigan, USA\\
\\ {\tt navo@nlp.cic.ipn.mx}, {\tt soujanya.poria@gmail.com},\\ {\tt gangeshwark@ihpc.a-star.edu.sg},\\ {\tt nchhaya@adobe.com}, {\tt mihalcea@umich.edu}, {\tt gelbukh@gelbukh.com},\\
}

\begin{document}
\maketitle

\begin{abstract}

Multimodal fusion is considered a key step in multimodal 
tasks such as sentiment analysis, emotion detection, question answering, and others. Most of the recent work on
multimodal fusion does not guarantee the fidelity of the multimodal representation
with respect to the unimodal representations. In this paper, we propose a variational
autoencoder-based approach for modality fusion that minimizes information loss between
unimodal and multimodal representations. We empirically show that this method outperforms
the state-of-the-art methods by a significant margin on several popular datasets.

\end{abstract}

\section{Introduction}
\label{sec:introduction}

Multimodal sentiment analysis has received significant traction in recent years, due to
its ability to understand the opinions expressed in the increasing number of videos available on  open platforms such as YouTube,
Facebook, Vimeo, and others. This is important, as more and more enterprises tend to make business
decisions based on the user sentiment behind their products as  expressed through these videos.

Multimodal fusion is considered a key step in multimodal sentiment analysis. Most recent work
on multimodal fusion \cite{poria-EtAl:2017:Long,AAAI1817390} has focused on the strategy of obtaining a multimodal   representation from the independent unimodal representations. Our approach takes this strategy one
step further, by also requiring that the original unimodal representations be reconstructed from the unified multimodal
representation. The motivation behind this is the intuition that different modalities are an expression
of the state of the mind. Hence, if we assume that the fused representation is the mind-state/sentiment/emotion, 
then in our approach we are ensuring that the fused representation can be mapped back to the unimodal
representations, which should improve the quality of the multimodal representation. In this paper,
we empirically argue that this is the case by showing that this approach outperforms the state-of-the-art in multimodal fusion.

We employ a variational autoencoder (VAE)~\cite{DBLP:journals/corr/KingmaW13},
where the encoder network
generates a latent representation from the unimodal representations. Further, the decoder network decodes
the unimodal representations from the latent representation to the original unimodal representation. This
latent representation is treated as the multimodal representation for the final classification.

%The rest of the paper is organized as follows: \cref{sec:related-works} briefs on the recent works, 
%\cref{sec:method} discusses our approach, \cref{sec:experiments} states the experimental settings,
%\cref{sec:results-discussion} reports the outcome of our experiments and its implications, and finally 
%\cref{sec:conclusion} makes a concluding remark.

\section{Related Work}
\label{sec:related-works}

\citet{rozgic2012ensemble} and \citet{wollmer2013youtube} were the first to fuse acoustic, visual, and
text modalities for sentiment and emotion detection. Later, \citet{poria2015deep} employed CNN and
multi-kernel learning for multimodal sentiment analysis. Further, \citet{poria-EtAl:2017:Long} used
long short-term memory (LSTM) to enable context-dependent multimodal fusion, where the surrounding
utterances are taken into account for context.

Recently, for context-free setting where the surrounding utterances are not used as context,  \citet{zadeh-EtAl:2017:EMNLP2017} used tensor outer-products to model
intra- and inter-modal
interactions. Again, \citet{AAAI1817341} used multi-view learning for utterance-level multimodal fusion.
Further, \citet{AAAI1817390} employed hybrid LSTM memory components to model intra-modal and cross-modal
interactions.
%Recently, \citet{bagherzadeh-EtAl:2018:Long} proposed an hierarchical approach for multimodal
%fusion that yields results comparable to \citet{AAAI1817341}.

\section{Method}
\label{sec:method}

Usually humans express their thoughts through three perceivable modalities -
textual (speech), acoustic (pitch and other properties of voice), and visual
(facial expression). Where most recent works on multimodal fusion treat these
unimodal representations independently and employ an encoder network
to obtain the fused representation vector, we go one step further by decoding the
fused-multimodal representation into the original unimodal representations.

\begin{figure}
    \centering
    \includegraphics[trim={5cm 4cm 9cm 5cm},width=\linewidth]{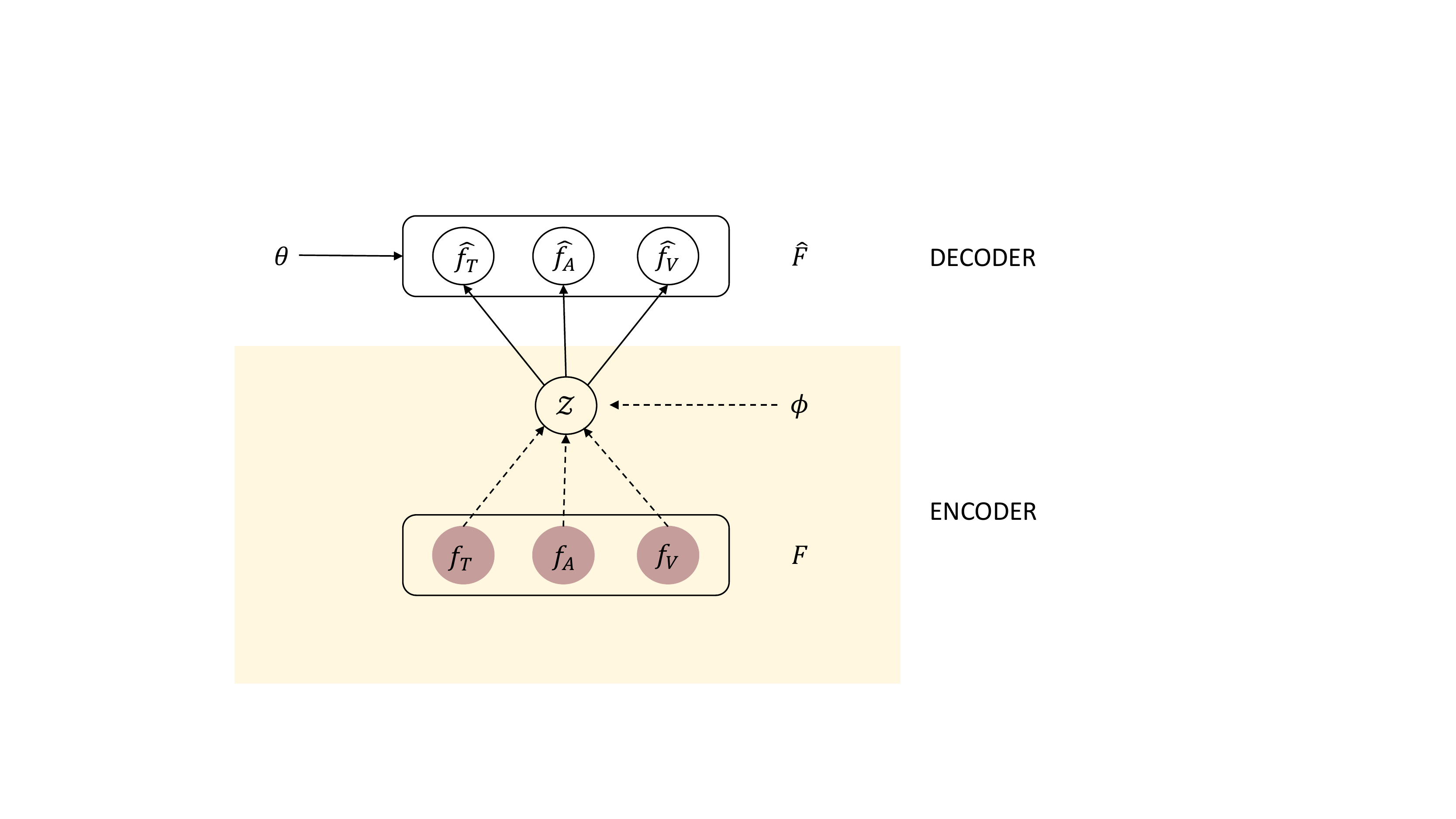}
    \caption{Graphical model of our multimodal fusion scheme.}
    \label{fig:model}
\end{figure}

First the utterance-level unimodal features are extracted independently. Then, the modality
features are fed to encoder network to sample the fused representation. Further,
the fused representation is decoded back to the unimodal representations to
ensure the fidelity of the fused representation. This setup is basically an
autoencoder. Specifically, we employ a variational autoencoder (VAE)~\cite{DBLP:journals/corr/KingmaW13},
as described in \cref{fig:model},
where the latent representation is used as the fused representation.

\subsection{Unimodal Feature Extraction}
\label{sec:unim-feat-extr}

Textual ($f_t$), visual ($f_v$), and acoustic ($f_a$) features are extracted using CNN, 3D-CNN~\cite{tran2015learning},
and OpenSmile~\cite{eyben2015opensmile} respectively, with the methodology described by
\citet{poria-EtAl:2017:Long}.

\subsection{Encoder}
\label{sec:encoder}

The encoder takes the concatenation of the unimodal features $F$ of an utterance as input, where
$f_t$ is textual feature of size $D_t$, $f_a$ is acoustic feature of size $D_a$,
and $f_v$ is visual feature of size $D_v$, and infers the latent multimodal representation
$z$ of size $D_z$ from the posterior distribution $p_\theta(z|F)$, such that
\begin{flalign}
    F&=f_t\oplus f_a\oplus f_v,\\
    p(F)&=\int p_\theta(F|z)p(z) dz,\\
    p(z)&=\mathcal{N}(0,I).
\end{flalign}
Since, the true posterior distribution $p_\theta(z|F)$ is intractable, $F$ is fed
through two fully-connected layers to generate mean ($\mu_{enc}$) and standard
deviation ($\sigma_{enc}$) of the approximate posterior normal distribution
$q_\phi(z|F)=\mathcal{N}(\mu_{enc},\sigma_{enc})$, which infers the latent
representation $z$:
\begin{flalign}
    h_1&=\ReLU(W_{h_1}F+b_{h_1}),\\
    \mu_{enc}&= W_{\mu}h_1+b_{\mu},\\
    \sigma_{enc}&=\softplus(W_{\sigma}h_1+b_{\sigma}),
\end{flalign}
where 
%$f_t\in \mathbb{R}^{D_t}$, $f_a\in \mathbb{R}^{D_a}$, $f_v\in
%\mathbb{R}^{D_v}$,
$F\in \mathbb{R}^{D_t+D_a+D_v}$, $W_{h_1}\in
\mathbb{R}^{D_{h}\times (D_t+D_a+D_v)}$, $b_{h_1}\in \mathbb{R}^{D_{h}}$,
$h_1\in \mathbb{R}^{D_{h}}$, $W_{\{\mu,\sigma\}}\in \mathbb{R}^{D_z\times D_{h}}$,
$b_{\{\mu,\sigma\}}\in \mathbb{R}^{D_z}$, $\mu_{enc}\in \mathbb{R}^{D_z}$, and
$\sigma_{enc}\in \mathbb{R}^{D_z}$.

%A more sophisticated encoder, like TFN~\cite{zadeh-EtAl:2017:EMNLP2017} and MFN~\cite{AAAI1817341}, can be used. However, encoding the unimodal representations is not the focus of this paper, rather it is the reconstruction of the unimodal features from the multimodal representation.

\paragraph{Sampling Latent (Multimodal) Representation}

The latent representation $z\sim q_\phi(z|F)$ is sampled using the
reparameterization trick~\cite{DBLP:journals/corr/KingmaW13} to facilitate backpropagation:
\begin{flalign}
    z&=\mu_{enc}+\epsilon \odot \sigma_{enc},\\
    \epsilon&\sim \mathcal{N}(0,I),
\end{flalign}
where $z\in \mathbb{R}^{D_z}$, $\epsilon\in \mathbb{R}^{D_z}$, and $\odot$
represents hadamard product. This $z$ is considered as the fused
multimodal representation.

\subsection{Decoder}
\label{sec:decoder}

The decoder reconstructs the input as $\hat{F}$ from the latent representation $z$ 
with two fully-connected layers as follows:
\begin{flalign}
    h_3&=\softplus(W_{h_3}z+b_{h_3}),\\
    \hat{F}&=W_{rec}h_3+b_{rec},
\end{flalign}
where $W_{h_3}\in \mathbb{R}^{D_h\times D_z}$, $b_{h_3}\in \mathbb{R}^{D_h}$,
$W_{rec}\in \mathbb{R}^{(D_t+D_a+D_v)\times D_h}$, $b_{rec}\in
\mathbb{R}^{(D_t+D_a+D_v)}$, $h_3\in \mathbb{R}^{D_h}$, and $\hat{F}\in
\mathbb{R}^{(D_t+D_a+D_v)}$.

%Similar to the encoder network, decoder network construction is not the focus of this paper. A more sophisticated decoder can be used.

\subsection{Classification}
\label{sec:classification}

We tried two different classification networks:

\paragraph{Logistic Regression (LR)}
We employ a fully-connected layer with softmax activation where the fused representation $z$ is fed:
\begin{flalign}
    \label{eq:SC}
    \mathcal{P}&=\softmax(W_{cls}z+b_{cls}),\\
    \hat{y}&=\argmax_i\mathcal{P}[i],
\end{flalign}
where $W_{cls}\in \mathbb{R}^{C\times D_z}$, $b_{cls}\in \mathbb{R}^{C}$,
$\mathcal{P}\in \mathbb{R}^{C}$ is the vector of class-probabilities,
$\hat{y}$ is the predicted class, and $C$ is the number of classes ($C=2$ in our case).

\paragraph{Context-Dependent Classifier (bc-LSTM~\cite{poria-EtAl:2017:Long})}
The sequence of fused utterance representations ($z_i$) in a video is fed to a
bidirectional-LSTM~\cite{hochreiter1997long}, following \citet{poria-EtAl:2017:Long},
of size $D_l$ for context propagation
and then the output of the LSTM is fed to a fully-connected layer with softmax activation for
classification:
\begin{flalign}
    \label{eq:CDC}
    Z&=[z_1, z_2,\dots,z_n],\\
    H&=\text{bi-LSTM}(Z),\\
    H&=[h_1, h_2,\dots,h_n],\\
    \mathcal{P}_j&=\softmax(W_{cls}h_j+b_{cls}),\\
    \hat{y}_j&=\argmax_i\mathcal{P}_j[i],
\end{flalign}
where $Z$ is the sequence of fused utterance representations in a video with $n$ utterances,
$H$ is the context-dependent fused representations of the utterances ($h_i\in \mathbb{R}^{2D_l}$),
$W_{cls}\in \mathbb{R}^{C\times 2D_l}$, $b_{cls}\in \mathbb{R}^{C}$,
$\mathcal{P}_j\in \mathbb{R}^{C}$ is the vector of class-probabilities for utterance $j$,
$\hat{y}_j$ is the predicted class for utterance $j$, and $C$ is the number of
classes (e.g. $C=2$ for MOSI dataset~(\cref{sec:datasets})).

\subsection{Training}
\label{sec:training}

\paragraph{Latent Representation Inference}

Following \citet{DBLP:journals/corr/KingmaW13}, the approximate posterior distribution $q_\phi(z|F)$ is tuned
close to the true posterior $p_\theta(z|F)$ by maximizing the evidence lower bound
(ELBO), where
\begin{flalign}
    \log p(F)&\ge\text{ELBO},\\
    \text{ELBO}&=\mathbb{E}_{q_\phi(z|F)}[\log p_\theta(F|z)]\nonumber\\
    &~~~~~-\KL[q_\phi(z|F)||p(z)].
\end{flalign}
\normalsize
The first term of the ELBO, $\mathbb{E}_{q_\phi(z|F)}[\log p_\theta(F|z)]$, corresponds to
the reconstruction loss of input $F$. The second term, $\KL[q_\phi(z|F)||p(z)]$,
pushes the approximate posterior $q_\phi(z|F)$ close to the prior
$p(z)=\mathcal{N}(0,I)$ by minimizing the KL-divergence between them.

\paragraph{Classification}

To train the sentiment classifier (\cref{sec:classification}), we minimize the categorical cross-entropy ($E$),
defined as
\begin{flalign}
    E=-\frac{1}{N}\sum_{i=1}^{N}\log\mathcal{P}_i[y_i],
\end{flalign}
where $N$ is the number of samples, $\mathcal{P}_i$ is the probability distribution for sample $i$
on different classes (for our experiments, we use two classes; positive and negative), and $y_i$ is the target
class for sample $i$.

The networks are optimized using stochastic-gradient-descent-based
Adam~\cite{DBLP:journals/corr/KingmaB14} 
algorithm. Further, the hyperparameters $\{D_h,D_l\}$ and learning-rate are
optimized with grid-search (optimal hyperparameters are listed in the
supplementary material). The latent
representation size $D_z$ is set to $100$.

\section{Experimental Settings}
\label{sec:experiments}

We evaluate the quality of the multimodal features extracted by VAE\footnote{implementation available at \url{https://github.com/xxxx/xxxx/} (will be releaved upon acceptance)} using two classifiers
(\cref{sec:classification}). Hence, the two variants are named VAE+LR and VAE+bc-LSTM in
\cref{tab:performance}.

\subsection{Datasets}
\label{sec:datasets}

We evaluate our approach on three different datasets.

\paragraph{MOSI~\cite{zadeh2016multimodal}}
This dataset contains videos of 89 people reviewing various
topics in English. The videos are segmented into utterances where each utterance is
annotated with sentiment tags
(\textit{positive/negative}). Our train/test splits of the dataset are completely disjoint with
respect to speakers. In particular, 1447 and 752 utterances are used for training
and test respectively.

\paragraph{MOSEI~\cite{bagherzadeh-EtAl:2018:Long}}
MOSEI dataset contains 22676 utterances from 3229 videos. The videos were crawled from Youtube. There are 1000 unique speakers in the MOSEI dataset. Videos in MOSEI mostly comprise of product and movie reviews. We used 16188, 1874, and 4614 utterances as training, validation, and test folds. respectively. The utterances are labeled with either of the \emph{positive, negative, and neutral} sentiment categories.

\paragraph{IEMOCAP~\cite{iemocap}}
IEMOCAP contains two way conversations
among ten speakers, segmented into utterances. The utterances are tagged with
one of the six emotion labels \textit{anger, happy, sad, neutral, excited, and frustrated}.
The first eight speakers of sessions one to four belong to training set and the rest to
the test set.
\begin{table}[ht]
    \centering
    \begin{tabular}{l|c|c}
        \hline
         Dataset & Train & Test\\
         \hline
         \hline
         MOSEI & 16188 & 4614 \\
         IEMOCAP & 5810 & 1623 \\
         MOSI & 1447 & 752 \\
         \hline
    \end{tabular}
    \caption{Utterance count in the train and test sets.}
    \label{tab:dataset}
\end{table}

% \paragraph{AVEC} not enough space; supplementary perhaps

%AVEC~\cite{Schuller:2012:ACA:2388676.2388776} dataset is a modification of
%SEMAINE database~\cite{5959155} containing interactions between humans and
%artificially intelligent agents. Each utterance of a dialogue is annotated with
%four real valued affective attributes: valence ($[-1,1]$), arousal ($[-1,1]$),
%expectancy ($[-1,1]$), and power ($[0,\infty)$). The annotations are available
%every 0.2 seconds in the original database. However, in order to adapt the
%annotations to our need of utterance-level annotation, we averaged the
%attributes over the span of an utterance.

% \paragraph{OMG} not enough space; supplementary perhaps

\subsection{Baseline Methods}

\paragraph{Logistic Regression (LR)}

The concatenation of the utterance-level unimodal representations is classified using
logistic regression as described in \cref{sec:classification}. This does not consider
the surrounding neighbouring utterances as context.

% \paragraph{Autoencoder (AE)}

% Instead of using VAE for multimodal fusion, regular autoencoder is used.

\paragraph{bc-LSTM~\cite{poria-EtAl:2017:Long}}

The concatenation of the utterance-level unimodal representations is sequentially fed to
the bc-LSTM classifier described in \cref{sec:classification}. This is the 
state-of-the-art method.

\paragraph{TFN~\cite{zadeh-EtAl:2017:EMNLP2017}}

This network models both intra-modal and inter-modal interactions through outer product.
It does not use the neighbouring utterances as context.

\paragraph{MFN~\cite{AAAI1817341}}

This network exploits multi-view learning to fuse modalities. It also does not
use neighbouring utterances as context.

\paragraph{MARN~\cite{AAAI1817390}}

In this model the intra-modal and cross-modal interactions are modeled with hybrid
LSTM memory component.

% \subsection{Model Variants}

% \paragraph{Variational Utterance-Level Classification (V+LR)}

% \paragraph{Variational Utterance-Level Contextual Classification (V+bc-LSTM)}

%\paragraph{Utterance-Level Multitask (ULM)} not enough space; supplementary perhaps

%\paragraph{Contextual Multitask (CM)} not enough space; supplementary perhaps

\section{Results and Discussion}
\label{sec:results-discussion}

\begin{table}[ht]
    \centering
    \begin{tabular}{l|c|c|c}
        \hline
         Method & MOSI & MOSEI & IEMOCAP \\
         \hline
         \hline
         TFN & 74.8  & 53.7 & 56.8 \\
         MARN & 74.5 & 53.2 &54.2\\
         MFN &74.2  & 54.1 & 53.5\\
         \hline
        %  AE+\hl{?} & \\
         LR & 74.6 & 56.6 & 53.9\\
         VAE+LR & 77.8 & 57.4 & 54.4\\
         \hline
         bc-LSTM & 75.1 & 56.8 & 57.7\\
         VAE+bc-LSTM & {\bf 80.4$^*$} & {\bf 58.8$^*$} & {\bf 59.6$^*$}\\
         \hline
    \end{tabular}
    \caption{Trimodal (acoustic, visual, and textual) F1-scores of our method against the baselines 
    (results on MOSI and IEMOCAP are based on the dataset split from \citet{poria-EtAl:2017:Long}); * signifies statistically significant improvement ($p<0.05$ with paired t-test) over bc-LSTM.}
    \label{tab:performance}
\end{table}

\cref{tab:performance} shows the performance our VAE-based methods, namely VAE+LR and VAE+bc-LSTM, outperform
their concatenation fusion counterpart LR and bc-LSTM consistently on all three datasets. Specifically,
our context-dependent model, VAE+bc-LSTM, outperforms the context-dependent state-of-the-art method bc-LSTM on
all the datasets, by 3.1\% on average.
Moreover, our context-free model VAE+LR outperforms the other context-free models, namely MFN, MARN, TFN, and LR, on all datasets, by 1.5\% on average. Also, due to the contextual information, VAE+bc-LSTM outperforms VAE+LR
by 3.1\% on average.

This is due to the superior multimodal representation from VAE, that retains enough information from the unimodal representations to allow reconstruction. This leads to highly informative classification. (Supplementary material compares the visualizations of the fused representations)

% \subsection{VAE vs. AE Fusion}

% \begin{figure}
%     \centering
%     \begin{subfigure}{0.48\linewidth}
%     \includegraphics[trim={2cm 1.5cm 0.2cm 1.8cm},width=\linewidth]{fig/VAE_train_iemocap.png}
%     \caption{}
%     \end{subfigure}
%     ~
%     \begin{subfigure}{0.48\linewidth}
%     \includegraphics[trim={2cm 1.5cm 0.2cm 1.8cm},width=\linewidth]{fig/AE_train_iemocap.png}
%     \caption{}
%     \end{subfigure}
%     \caption{(a) and (b) depict t-SNE scatter-plots of VAE and AE multimodal features respectively, for IEMOCAP.}
%     \label{fig:vae_vs_ae}
% \end{figure}

\subsection{Case Study}

Comparing the predictions of our model to the baselines reveals that our model is
better equipped for catching the instances where the non-verbal cues are essential for classification.
For instance, the utterance from IEMOCAP ``\textit{I still can't live on in six seven and five. It's not possible in Los
Angeles. Housing is too expensive.}'' is mis-classified as \textit{excited} by bc-LSTM, whereas VAE+bc-LSTM
correctly classifies it as \textit{angry}. We posit that in this case the bc-LSTM is confused by the
emotionally ambiguous textual modality, whereas the VAE+bc-LSTM taps into the visual modality to observe the
frown on the speakers face to make the correct classification. Besides this, we observed several similar
cases where VAE+bc-LSTM or VAE+LR correctly classifies based on non-verbal cues, where their non-VAE counterparts could not.

\paragraph{Error Analysis}

\textit{``No. I am just making myself fascinating for you.''}
is response to a question \textit{``you going out somewhere, dear?''}.
This is a sarcastic response. VAE+bc-LSTM falsely predicted the emotion as 
\emph{excited}, where the ground truth is \emph{angry}. We suspect that our model's
failure to identify sarcasm with the aid of multimodality led to this
misclassification.

\section{Conclusion}
\label{sec:conclusion}

In this paper, we  presented a comprehensive fusion strategy, based on VAE that outperforms previous
methods by a significant margin. The encoder and decoder networks in the VAE are simple fully-connected layers.
We plan to improve the performance of our method by employing more sophisticated networks, such as fusion networks like MFN and TFN as the encoders.

\bibliographystyle{acl_natbib}
\bibliography{bibexport_new.bib}
\end{document}